\documentclass[10pt,twocolumn,letterpaper]{article}
          
\usepackage[pagenumbers]{iccv} 

\usepackage{comment}
\usepackage{multirow}
\usepackage{float}
\usepackage{caption}
\usepackage{amsmath}
\usepackage{soul}
\usepackage{xcolor,pifont}
\usepackage[accsupp]{axessibility}

\renewcommand{\thefootnote}{\fnsymbol{footnote}}

\usepackage{lipsum}

\newcommand\blfootnote[1]{%
  \begingroup
  \renewcommand\thefootnote{}\footnote{#1}%
  \addtocounter{footnote}{-1}%
  \endgroup
}

\newcommand*\colourcheck[1]{%
  \expandafter\newcommand\csname #1check\endcsname{\textcolor{#1}{\ding{52}}}%
}

\newcommand*\colourcross[1]{%
  \expandafter\newcommand\csname #1cross\endcsname{\textcolor{#1}{\ding{55}}}%
}

\colourcheck{green}
\colourcheck{pink}
\colourcheck{cyan}
\colourcross{red}

\definecolor{corrcol}{rgb}{0.0, 0.0, 0.0}

\newcommand{\EoA}{$\hspace{170pt}$}

\definecolor{iccvblue}{rgb}{0.21,0.49,0.74}
\usepackage[pagebackref,breaklinks,colorlinks,allcolors=iccvblue]{hyperref}

\def\paperID{15063} 
\def\confName{ICCV}
\def\confYear{2025}

\title{\textit{HumanOLAT}: A Large-Scale Dataset for Full-Body Human Relighting and Novel-View Synthesis} 

\author{
Timo Teufel$^{1,*}$\quad\quad\quad
Pulkit Gera$^{1,*}$\quad\quad\quad
Xilong Zhou$^{1}$\quad\quad\quad
Umar Iqbal$^{2}$\\
\quad\quad
Pramod Rao$^{1}$\quad\quad\quad\;\;
Jan Kautz$^{2}$\quad\quad\;
Vladislav Golyanik$^{1}$\;
Christian Theobalt$^{1}$\\
$^{1}$Max Planck Institute for Informatics, SIC $\quad$ $^{2}$NVIDIA
}

\begin{document}
\twocolumn[{%
\renewcommand\twocolumn[1][]{#1}%
\maketitle
\begin{center}
    \vspace{-7mm}
    \centering
    \includegraphics[trim=1.5cm 0cm 1.4cm 0cm, clip,width=1.0\textwidth]{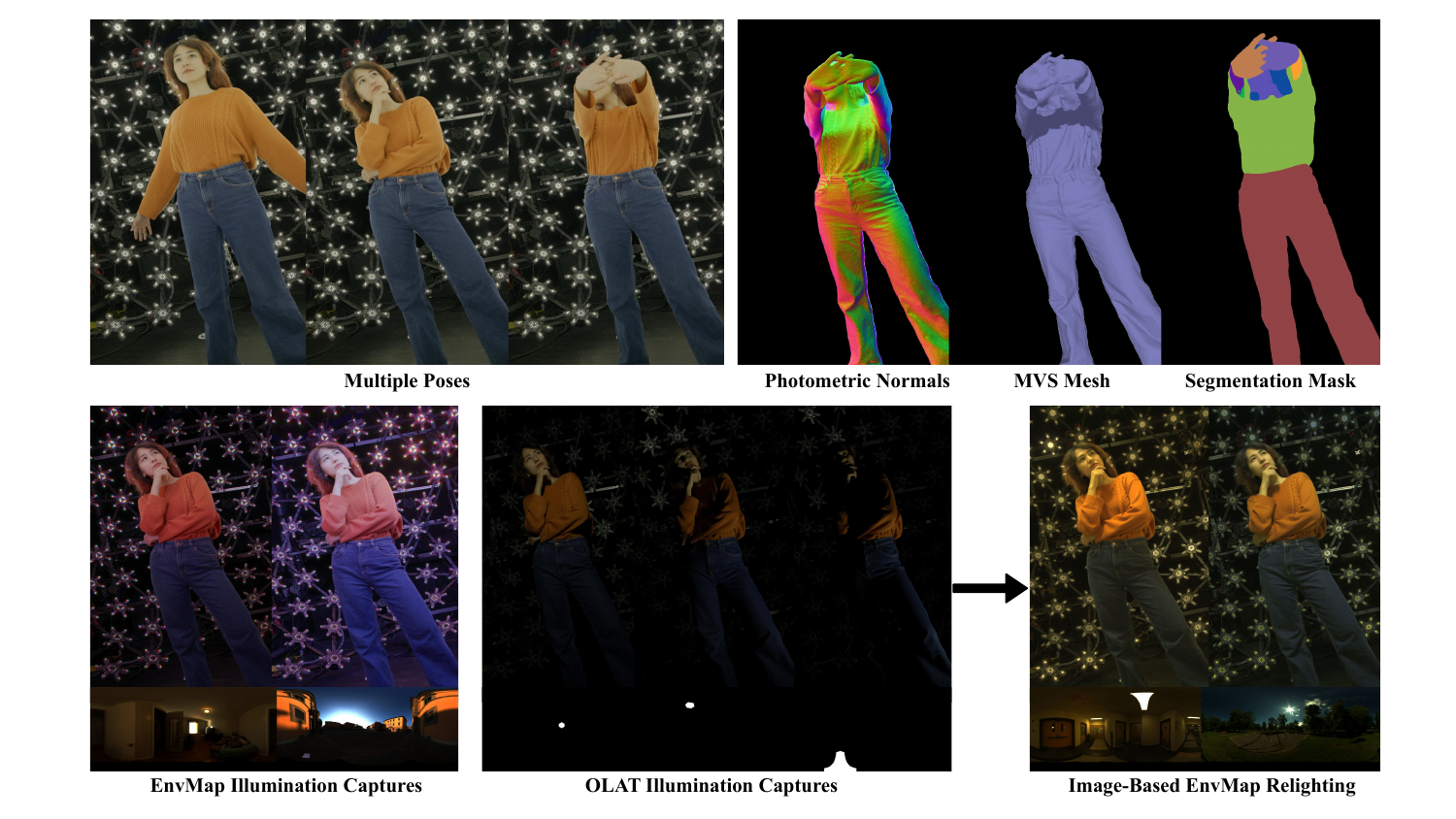}
         \vspace{-25pt}
    \captionof{figure}{\textbf{Overview of the proposed \textit{HumanOLAT} dataset.} We record $21$ subjects in three static poses each (top left) and provide pixel-wise photometric normals, multi-view stereo (MVS) meshes and segmentation masks (top right). To enable broad-scale application in various relighting tasks (e.g.~illumination harmonization), we capture every pose both under ten environment map illuminations loaded directly into our lightstage (bottom left) as well as $331$ OLAT illuminations suitable for image-based relighting~\cite{debevec2000acquiring} (bottom right).}
    
    \label{fig:data overview}
\end{center}
}]
\maketitle
\blfootnote{$\boldsymbol{^\star}$ both authors contributed equally to this work} 
\begin{abstract} 

\vspace{-5mm}

Simultaneous relighting and novel-view rendering of digital human representations is an important yet challenging task with numerous applications. Progress in this area has been significantly limited due to the lack of publicly available, high-quality datasets, especially for full-body human captures. To address this critical gap, we introduce the \textit{HumanOLAT} dataset, the first publicly accessible large-scale dataset of multi-view One-Light-at-a-Time (OLAT) captures of full-body humans. The dataset includes HDR RGB frames under various illuminations, such as white light, environment maps, color gradients and fine-grained OLAT illuminations. Our evaluations of state-of-the-art relighting and novel-view synthesis methods underscore both the dataset's value and the significant challenges still present in modeling complex human-centric appearance and lighting interactions. We believe \textit{HumanOLAT} will significantly facilitate future research, enabling rigorous benchmarking and advancements in both general and human-specific relighting and rendering techniques.

\end{abstract}

\section{Introduction}

Simultaneous relighting and novel-view rendering of digital human representations is a fundamental and long-standing challenge in computer vision and graphics, with applications spanning game and movie production, virtual telepresence, augmented reality (AR), virtual reality (VR), and mixed reality. Despite significant progress, the problem remains challenging due to inherent ambiguities in human appearance, arising from complex factors such as diverse clothing materials, intricate shadowing effects, and subsurface scattering in human skin.

Historically, the de facto standard for relighting research has been image-based relighting from One-Light-at-a-Time (OLAT) captures~\cite{debevec2000acquiring}, which effectively disambiguates appearance components, enabling accurate novel illumination synthesis. However, prior research has primarily focused on upper-body and facial relighting due to the complexities and practical challenges associated with full-body captures. Hence, there are only a handful of works that address this challenge. A critical obstacle in this domain has been the lack of publicly accessible and comprehensive full-body OLAT datasets.

Several factors contribute to the scarcity of these datasets. Firstly, acquiring OLAT data requires specialized, costly and not widely accessible hardware setups known as lightstages, enabling precise control over illumination and multi-view camera setups. Secondly, capturing full-body OLAT data is considerably challenging due to the necessity for larger capturing space and extended capture durations, during which even minor involuntary movements of subjects result in noticeable artifacts. Consequently, a large number of existing methods rely on synthetic data~\cite{iqbal2022ranarelightablearticulatedneural}, learned priors or regularizers~\cite{chen2022relighting,Luvizon2024relightneuralactor}. These methods, however, often exhibit diminished rendering quality with visible appearance artifacts.

To address the notable absence of a publicly available dataset for this task, this work introduces  \textit{HumanOLAT}. Our full-body human OLAT dataset consists of 21 diverse subjects, each captured in three distinct poses, providing a comprehensive resource for the community. The dataset includes precise camera and illumination calibrations, multi-view RGB frames under color-gradient illumination for photometric normal estimation, fine-grained OLAT captures, and images under predefined environment maps. \textit{HumanOLAT} facilitates multiple evaluation scenarios, including static full-body novel-view synthesis, static full-body relighting, and joint full-body novel-view synthesis and lighting. An illustration of the captures and annotations from \textit{HumanOLAT} is depicted in Fig.~\ref{fig:data overview}. 

We perform extensive experiments using several state-of-the-art relighting and novel-view synthesis methods \cite{zhang2024prtgaussian, bi2024gs3, Fan2025RNG, zhenyuan2024bigs} to demonstrate the utility and challenges posed by our dataset. Our evaluations highlight key remaining challenges in the field, pinpointing opportunities for further methodological improvements.  

To summarise, the main contribution of this paper is \textit{HumanOLAT}, a new OLAT multi-illumination dataset providing physically-based ground truth under any lighting condition for addressing different problems in the area of human relighting, novel view synthesis, and potentially other tasks in future. \textit{HumanOLAT} is publicly available at \href{https://vcai.mpi-inf.mpg.de/projects/HumanOLAT/}{https://vcai.mpi-inf.mpg.de/projects/HumanOLAT/}.
\section{Related Work}\label{sec:RW}

\subsection{Lightstages}

A light stage is an active illumination system allowing fine-grained control over lighting. First introduced by Debevec et al.~\cite{debevec2000acquiring} to acquire the reflectance field of human faces, these capture systems enable the collection of detailed multi-view image data with known illumination and as such, have become foundational in for developing and benchmarking realistic relighting methods \cite{kim2024switchlight, pandey2021total, saito2024rgca, li2024uravatar, chen2024urhand, guo2019relightables, zhou2023relightable}. Crucially, compared to other setups collecting real-world multi-illumination data \cite{Luvizon2024relightneuralactor}, lightstage's ability to precisely control lighting allows for the capture of detailed reflectance data in the form of one-light-at-a-time (OLAT) images. As light transport is linear \cite{debevec2000acquiring}, these illuminations can be additively combined to enable physically correct image-based relighting under arbitrary target light.

While a variety of works provide lightstage OLAT data for static objects \cite{toschi2023relightnerfdatasetnovel, liu2023openillumination}, faces \cite{stratou2011effectofillumination, zhang2021neuralvideoportraitrelighting, saito2024rgca} and hands\cite{chen2024urhand}, to our knowledge, no publicly available OLAT data for full-body humans exists. With our proposed \textit{HumanOLAT} dataset, we aim to close this gap by providing varied multi-illumination data,  including OLATs, for a large set of diverse humans.

\subsection{Human Relighting}

\paragraph{General Object-centric Relighting}

Recent works exploring the relighting of static objects have largely focused on neural rendering approaches. Early works \cite{zhang2021nerfactor, boss2021nerd, srinivasan2021nerv} extend Neural Radiance Fields (NeRF) \cite{nerf} to enable relighting by decomposing the scene into intrinsic components such as surface normals, albedo, and roughness. While these methods achieve impressive results for simple objects, they struggle with complex materials and lighting effects commonly found in human subjects. Follow-up works \cite{physg2021} improve upon this by incorporating physically-based rendering principles, enabling more accurate modeling of specular reflections and global illumination effects.

Following the success of 3D Gaussian Splatting (3DGS) \cite{kerbl20233d}, numerous works have explored its application to relighting static, object-centric scenes \cite{gau2023R3DG, shi2023gir, liang2023gsir, du2024gsID, wu2024deferredgs, jiang2023gaussianshader, zhu2024gsror, zhang2024prtgaussian, zhenyuan2024bigs, kuang2024olat, bi2024gs3, yong2025omgopacitymattersmaterial, liang2024gusir, dihlmann2024subsurface}. These approaches extend the original splatting technique by parametrizing each 3D Gaussian primitive with both BRDF reflectance properties (albedo, roughness, specular) and standard Gaussian attributes (opacity, mean, scale) to enable lighting-aware rendering. The field has seen advances in both settings with unknown illumination \cite{gau2023R3DG, shi2023gir, chen2024gi, du2024gsID, wu2024deferredgs, jiang2023gaussianshader, zhu2024gsror, liang2023gsir} and known lighting conditions \cite{zhang2024prtgaussian, zhenyuan2024bigs, kuang2024olat, bi2024gs3, yong2025omgopacitymattersmaterial, dihlmann2024subsurface}.

For these methods, the human body presents significant relighting challenges due to its intricate geometry, diverse material properties (including skin, hair, and clothing), and complex lighting effects, such as subsurface scattering and self-shadowing. Our proposed dataset, therefore, serves as a valuable benchmark for evaluating and stress-testing these approaches under realistic conditions.
\vspace{-5mm}
\paragraph{Illumination Harmonization}

Given an arbitrary foreground input and a target lighting condition, such as a text prompt or a background image, illumination harmonization methods aim to generate photo-realistic depictions of the foreground under the new lighting. Recent state-of-the-art methods \cite{kanamori2018relightinghumans, pandey2021total, ji2022geometry, kim2024switchlight, He2024DifFRelight} utilize encoder-decoder architectures \cite{kanamori2018relightinghumans, pandey2021total, ji2022geometry, kim2024switchlight} or diffusion models \cite{He2024DifFRelight, zhang2025scaling} to directly learn image-based relighting from a set of ground truth data. The latest advancement, IC-Light by Zhang et al.~\cite{zhang2025scaling}, leverages multiple publicly available datasets \cite{liu2023openillumination, deitke2022objaverseuniverseannotated3d} and introduces an explicit light transport consistency loss to achieve plausible illumination harmonization. 

Recent works \cite{Sun2019SingleImage, pandey2021total, kim2024switchlight, He2024DifFRelight} have shown that OLAT data is particularly effective for training these methods, as it enables realistic relighting under arbitrary environment maps\cite{debevec2000acquiring}. With \textit{HumanOLAT}, we aim to provide such data for full-body captures and make it publicly accessible to the research community.
\vspace{-5mm}
\paragraph{Drivable and Relightable Human Avatars}

The scarcity of publicly available full-body multi-illumination data has driven numerous studies \cite{chen2024meshavatarlearninghighqualitytriangular, li2024animatable, xu2023relightableanimatableneuralavatar, lin2023relightableanimatableneuralavatars, Wang2024IntrinsicAvatar, Zheng2024PhysAavatar, iqbal2022ranarelightablearticulatedneural} to focus on creating human avatars using single-illumination datasets like ActorsHQ~\cite{iwase2023relightablehands} and MVHumanNet~\cite{Xiong2024MVHumanNet}. These methods typically employ a mix of inverse rendering to deduce underlying lighting conditions \cite{chen2024meshavatarlearninghighqualitytriangular, li2024animatable, xu2023relightableanimatableneuralavatar} or use synthetically generated multi-illumination datasets \cite{lin2023relightableanimatableneuralavatars, Wang2024IntrinsicAvatar, Zheng2024PhysAavatar, iqbal2022ranarelightablearticulatedneural} to learn the intrinsic properties essential for physically-based relighting. In contrast, Luvizon et al.~\cite{Luvizon2024relightneuralactor} capture dynamic subjects under four distinct lighting conditions using light probes to obtain HDRI environment maps. Although this approach provides real-world relightable captures, it lacks fine-grained control over lighting conditions, resulting in a noticeable quality gap compared to methods utilizing light stage-based datasets \cite{guo2019relightables, zhou2023relightable, WangARXIV2025fullbodyrgca}. Guo et al.~\cite{guo2019relightables} and Zhou et al.~\cite{zhou2023relightable} utilize images captured with white light and color gradients to directly estimate intrinsic properties, such as albedo and geometry, thereby enabling high-quality relighting of both animated and static human models. Notably, the concurrent work by Wang et al.~\cite{WangARXIV2025fullbodyrgca} introduces a method for creating drivable and relightable full-body avatars based on 3DGS \cite{kerbl20233d}, achieving exceptional quality by learning light transfer from dynamic full-body OLAT data. Similar studies focusing on relightable hand \cite{iwase2023relightablehands, chen2024urhand} and head models \cite{bi2021deeprelightable, yang2023towards, saito2024rgca, li2024uravatar} using OLAT data also demonstrate similarly high-quality results. 

With the \textit{HumanOLAT} dataset, we aim to further accelerate the development of high-quality relightable human avatars by offering fine-grained OLAT data of static humans for training and benchmarking.

\section{The \textit{HumanOLAT} Dataset}
\label{sec:HumanOLAT}

\begin{figure*}[h!]
\includegraphics[trim=2.3cm 1cm 2.3cm 0.5cm, clip, width=1.0\textwidth]{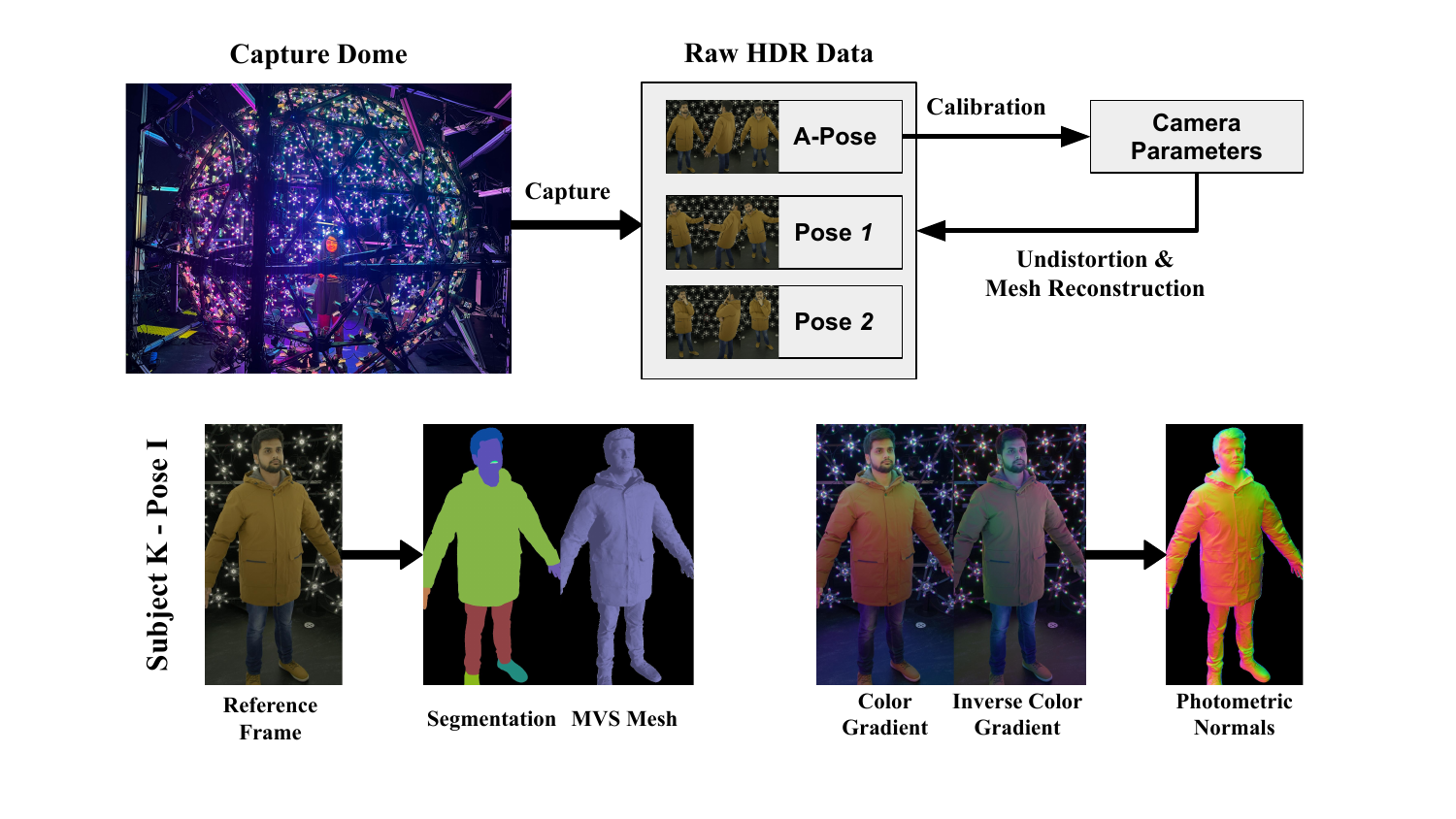}
\centering
\vspace{-25pt}

\caption{\textbf{Scheme of our data processing pipeline}. For each subject, we capture multi-view multi-illumination data under an A-pose as well as two creative poses. Afterwards, we use Metashape~\cite{agisoft_metashape} to establish per-subject camera parameters from the A-pose. Taking a frame illuminated under white light as reference, we construct a multi-view stereo (MVS) mesh for each captured pose using the same software and employ Sapiens~\cite{khirodkar2024sapiensfoundationhumanvision} for generating segmentation masks. Finally, we deduce pixel-wise photometric normals from color gradient illuminations~\cite{guo2019relightables, zhou2023relightable} recorded during the capture.}

\vspace{-0.2in}
\label{fig:processing_overview}
\end{figure*}

In the following, we describe the contents and the capture setup of the proposed dataset in detail. Specifically, we describe our capture setup in Sec.~\ref{subsec: setup} and provide an overview of the recorded data in Sec.~\ref{subsec: datasetoverview}, followed by an explanation of our data processing pipeline in Sec.~\ref{subsec: data processing}. Finally, we compare \textit{HumanOLAT} to existing publicly available lightstage datasets in Sec.~\ref{sec:compare_with_ls_datasets}.

\subsection{Lightstage Capture Setup}
\label{subsec: setup}

Fig.~\ref{fig:processing_overview} illustrates our capture setup, which features a spherical dome equipped with 40 RED Komodo 6K cameras and 331 individually controllable LEDs capable of emitting red, green, blue, amber, and white light (RGBAW). The cameras and lights are arranged $360^{\circ}$ around the subject, enabling the capture of synchronized multi-illumination sequences at 30 FPS with the 5K image resolution.  

\subsection{Dataset Description}
\label{subsec: datasetoverview}

\textit{HumanOLAT} is a comprehensive multi-view, multi-illumination dataset designed for full-body relighting. It includes $21$ subjects, each captured in three distinct poses: an A-pose and two creative poses. For each recording, we capture the subject's appearance from $40$ different views under the following diverse lighting conditions:\\
\vspace{-0.1in}

\begin{itemize}[align=parleft, labelsep=0.4cm]
    \item
    \textbf{One white light illumination} utilized for camera calibration, mesh reconstruction, segmentation, and providing ground truth for albedo~\cite{zhou2023relightable}; 
    \item
    \textbf{Two color gradient illuminations}~\cite{guo2019relightables, zhou2023relightable} employed to estimate pixel-wise photometric normals; 
    \item
    \textbf{Ten environment map illuminations}~\cite{guo2019relightables, zhou2023relightable} used directly by methods that assume a known environment map~\cite{zhenyuan2024bigs, Luvizon2024relightneuralactor}; 
    \item
    \textbf{331 OLAT Illuminations} applied in methods that rely on single light images~\cite{zhang2024prtgaussian, bi2024gs3, zhenyuan2024bigs} and enable image-based relighting under arbitrary target lighting~\cite{debevec2000acquiring}. 
\end{itemize}

\begin{figure}[h]
\includegraphics[trim=7.3cm 0.4cm 7.3cm 0.4cm, clip, width=0.48\textwidth]{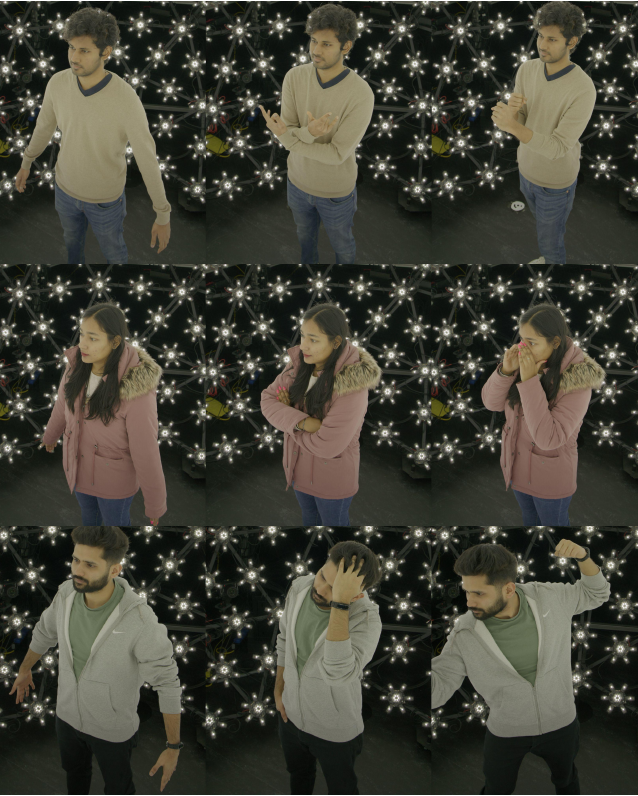}
\centering
\vspace{-0.2in}

\caption{Samples of captures of \textit{HumanOLAT}. The subjects are wearing a variety of clothing and take three different poses each.} 
\vspace{-0.2in} 

\label{fig:subject_mosaic}
\end{figure}

Samples of captured illuminations and subjects are shown in Fig.~\ref{fig:data overview} and Fig.~\ref{fig:subject_mosaic}. For each of the 21 subjects, the dataset includes approximately:
$3 \text{ Poses} \times 40 \text{ Views} \times 344 \text{ Illuminations} \approx \text{40K Frames}$ resulting in a total of around 850K frames.

As shown in Fig.~\ref{fig:subject_mosaic}, the subjects include both male and female individuals wearing a variety of clothing, ranging from tight-fitting shirts and jumpers to loose, volumetric clothing such as hoodies and jackets. Specifically, among the recorded subjects, there are $13$ males and $8$ females. Of these, $5$ subjects are wearing tight-fitting shirts and jumpers, while $16$ are wearing loose clothing, which includes $7$ jumpers, $5$ hoodies, and $3$ jackets. Please refer to \cref{fig:large_data_samples} in the supplement for additional data samples. 

We also provide OpenPose~\cite{cao2019openpose} annotations and SMPL-X~\cite{SMPL-X:2019} parameters. For more details, please refer to App.~\ref{app:openpose_smplx}. 

\subsection{Data Processing}
\label{subsec: data processing}

\subsubsection{Calibration and Mesh Reconstruction}

For estimating camera parameters and suitable initial geometry, we rely on feature-based algorithms implemented in the proprietary software Metashape~\cite{agisoft_metashape}. To ensure consistent camera intrinsics and extrinsics across different poses, we calculate a single camera calibration for each subject using the A-pose capture as a reference. Since the quality of feature-based calibration and reconstruction is directly dependent on the quantity and quality of features, we use the frame illuminated under flat white light to ensure optimal conditions for feature detection. Finally, we use a marker-based setup to transform estimated camera poses and geometry into a predefined real-world canonical coordinate system. The positions of individual lights are also provided within the same canonical system by measuring their physical location. However, we note that while most of the LEDs are static, a subset of around ${\sim}20$ lights are attached to a hatch used to enter the lightstage. Since the hatch moves, fully accurate 3D light positions cannot be guaranteed for this subset, and we recommend ignoring them during training and evaluation. Note, however, that methods that do not rely on accurate 3D light positions, such as ones performing relighting from environment maps, are unaffected.

\begin{figure}[h]
\includegraphics[trim=2.3cm 2cm 2.3cm 2cm, clip, width=0.48\textwidth]{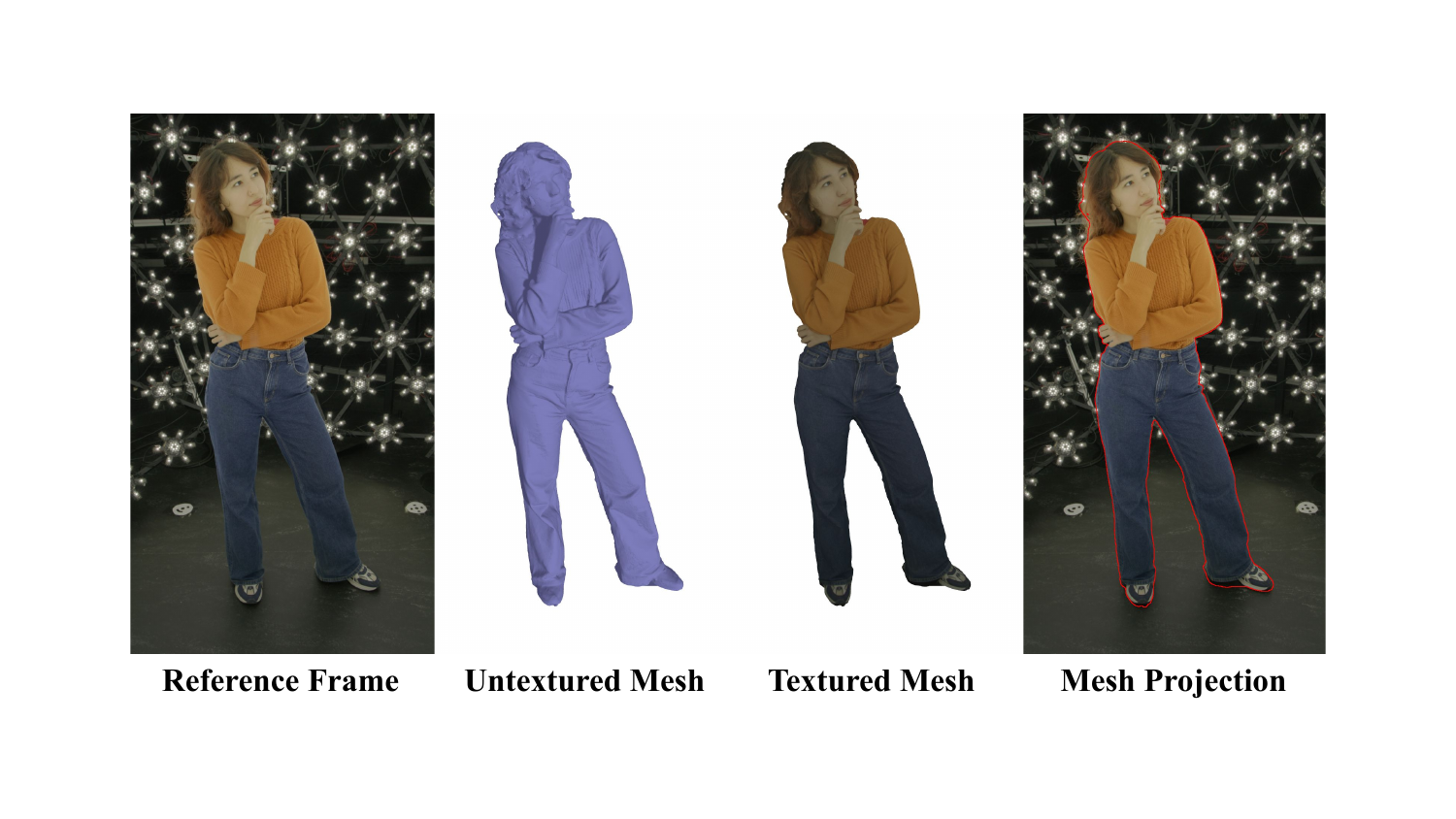}
\centering
\vspace{-0.2in}

\caption{Illustration of the MVS meshes provided within \textit{HumanOLAT}. From left to right: (1) one of the $40$ frames used for reconstruction, (2) the mesh rendered under flat color, (3) the mesh rendered with texture and (4) border of the mesh rendered onto the reference frame.} 
\label{fig:mesh_quality}
\vspace{-0.2in}

\end{figure}

To quantitatively validate our calibrations, we calculate the average re-projection error of triangulated keypoints across a representative subset of subjects in the dataset. In sum, we arrive at an average error of $0.819$ pixels. Additionally, we qualitatively assess the quality of the extracted mesh in Fig.~\ref{fig:mesh_quality} by projecting the extracted textured mesh onto the camera image plane using the calibrated camera parameters. It can be seen that the mesh aligns well with the reference frame, demonstrating the accuracy of camera calibration and the mesh extraction.

\subsubsection{Mask Generation}

Using estimated camera distortion parameters, we undistort captured images in accordance with the pin-hole camera model. Afterwards, to generate masks, we rely on the Sapiens models introduced by Khirodkar et al.~\cite{khirodkar2024sapiensfoundationhumanvision}.
We find that Sapiens, being designed for human-centric vision tasks, generally produces more accurate masks in our case compared to other popular methods, such as SAM \cite{kirillov2023segment, ravi2024sam2segmentimages}.

\subsubsection{Motion Compensation}
\label{subsec: motion compensation}

\begin{figure}[h]
\includegraphics[width=0.48\textwidth]{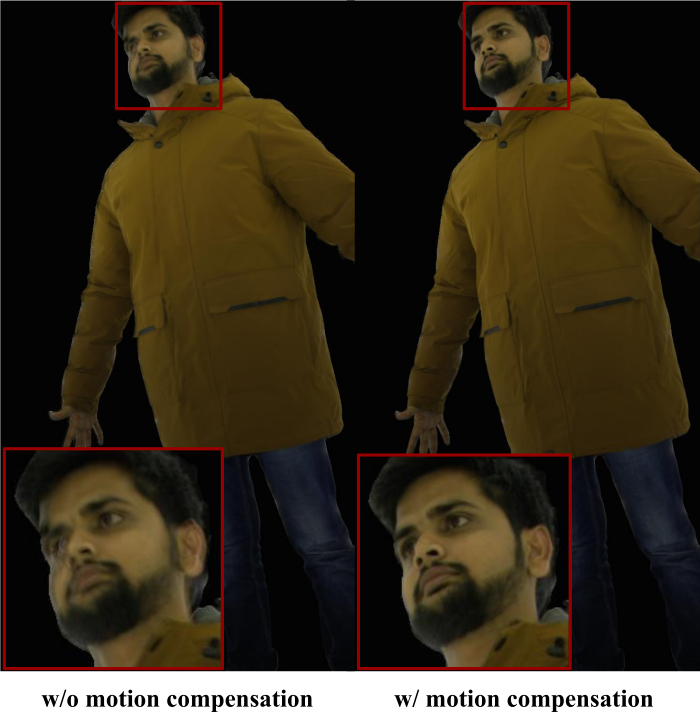}
\centering
\vspace{-0.2in}
\caption{Comparison of summing OLAT illumination without (left) and with (right) motion compensation applied.}
\vspace{-0.2in}

\label{fig:motion_compensation}
\end{figure}

We note that as a single recording takes around 11 seconds, the subjects are unable to remain fully static over the course of a single capture and sway slightly. As illustrated in Fig.~\ref{fig:motion_compensation}, for image-based relighting according to Debevec et al.~\cite{debevec2000acquiring}, this leads to blurry results unsuitable for training. To combat this, we follow the method by Wenger et al.~\cite{wenger2005performancerelighting} to perform motion compensation: injecting a white light frame every $21$th OLAT frame, we calculate optical flow towards a target frame---in our case the white illumination frame also used for mesh reconstruction and segmentation---using tracking-any-point (TAP) foundation model Co-Tracker3~\cite{karaev2024cotracker3} by tracking points sampled on the target frame over the concatenated tracking frames. Note that to keep computational cost reasonable, we track only ${\sim}12$k sparse grid points and linearly interpolate between them to arrive at the final dense flow. Finally, to motion-correct a given OLAT image, we linearly interpolate the flow for the two adjacent tracking frames and warp towards the target frame. See Fig.~\ref{fig:motion_compensation} for the qualitative results comparing summed raw and motion-compensated OLATs. 

\subsubsection{Normals and Image-Based Relighting} 

Following Guo et al.~\cite{guo2019relightables} and Zhou et al.~\cite{zhou2023relightable}, we provide pixel-wise photometric normals $\textbf{n}$ computed from color and inverse color gradient illuminations $g^{+}$ and $g^{-}$ as

\begin{equation}\label{eq:normals}
     \textbf{n} = \frac{\textbf{d}}{|\textbf{d}|},\quad \text{with}\quad\textbf{d} = \frac{g^{+} - g^{-}}{g^{+} + g^{-}}. 
\end{equation}

We leverage the linearity of lighting to achieve accurate image-based relighting under arbitrary environment maps. Specifically, given a target environment map $E_\text{target}$, we obtain the weighting color $\textbf{c}_{i}$ for each OLAT image by masking $E_\text{target}$ with each OLAT environment mask $E_{i}$ and subsequent per-channel averaging. According to Debevec et al.~\cite{debevec2000acquiring}, we then perform a per-channel linear combination of $N_\text{OLAT}$ of motion-corrected OLAT frames $I_{i}$ to acquire the final relit image $I_\text{target}$. This process is formulated as

\begin{equation}\label{eq:olat}
    I_\text{target} = \sum_{i=0}^{N_\text{OLAT}} \textbf{c}_{i} I_{i}. 
\end{equation}

We visualize the estimated photometric normals from \cref{eq:normals} and show relighting results obtained with \cref{eq:olat} in Figs.~\ref{fig:data overview} and \ref{fig:processing_overview}. 

\subsection{Comparison with Existing Lightstage Datasets}
\label{sec:compare_with_ls_datasets}

\begin{table*}[htb]
\begin{center}
\scalebox{0.96}{
\begin{tabular}{ c c c c c c c c c c }\toprule
 \multirow{2}{*}{Dataset} & \multirow{2}{*}{Type} & \multirow{2}{*}{\#Frames} & \multirow{2}{*}{\#Subj.} & \multirow{2}{*}{\#View} & \multirow{2}{*}{Res.} & \multicolumn{4}{c}{Illuminations}\\ 
 &&&&&&White& ColorGrad. & EnvMap & OLAT\\ 
 
 \midrule
  \multicolumn{1}{l}{ReNe \cite{toschi2023relightnerfdatasetnovel}} & objects & $40$k & $20$ & $50$ &$1$K & \redcross & \redcross & \redcross & \greencheck \\
  \multicolumn{1}{l}{OpenIllumination \cite{liu2023openillumination}$^{+}$} &objects & $108$k & $64$ & $72$  & $4$K  & \greencheck & \redcross & \redcross & \greencheck\\ \midrule
  \multicolumn{1}{l}{ICT-3DRFE \cite{ma2007rapidacquisition, stratou2011effectofillumination}$^\dagger$} & heads & $14$k & $23$ & $2$ & $1$K & \redcross & \cyancheck & \redcross & \redcross \\ 
  \multicolumn{1}{l}{Dynamic OLAT \cite{zhang2021neuralvideoportraitrelighting}$^{\star}$} & heads & $603$k & $4^{*}$ & $1$ &  $4$K & \greencheck & \redcross & \redcross & \cyancheck\\ 
  \multicolumn{1}{l}{Goliath-4~\cite{saito2024rgca, chen2024urhand}$^{\star}$} & hands/heads & ${>}1$M & $4$ & 150/110 &   $4$K  & \greencheck & \redcross & \redcross & \cyancheck \\\midrule
  \multicolumn{1}{l}{Ultrastage \cite{zhou2023relightable}} & full bodies & $192$k & $100$ & $32$ & $8$K & \greencheck & \greencheck & \redcross & \redcross \\
  \multicolumn{1}{l}{\textit{HumanOLAT}} & full bodies & $850$k & $21$ & $40$ & $5$K & \greencheck & \greencheck & \greencheck & \greencheck \\
 \bottomrule\\
\vspace{-0.3in}
\end{tabular}
}
  \caption{Comparison of \textit{HumanOLAT} with existing publicly available multi-illumination lightstage datasets regarding the number of frames (\#Frames), captured subjects (\#Subj.), views (\#View), resolution (Res.) and provided types of illuminations. For the latter, ``\greencheck''~and ``\redcross'''~denote whether a specific illumination is or is not contained within the dataset, while ``\cyancheck''~implies the illuminations are present in limited capacity. (``$^{+}$'': OpenIllumination provides $13$ pattern illuminations in addition to OLAT data; ``$^{\dagger}$'': ICT-3DRFE does not provide raw captures, but diffuse and specular normal and reflectance maps estimated from color gradients; ``$^{\star}$'': Dynamic OLAT and Goliath-4 provide \textit{grouped} OLAT illuminations---with a subset of the lights on---for dynamically moving subjects).}
   \label{tab:dataset_compare}
\end{center}
\vspace{-6mm}
\end{table*}

A comparison of \textit{HumanOLAT} with publicly available lightstage multi-illumination datasets is shown in \cref{tab:dataset_compare}. For object-centric scenes, ReNé \cite{toschi2023relightnerfdatasetnovel} and OpenIllumination \cite{liu2023openillumination} 
both capture multi-view OLAT images under a variety of light positions, with the latter also providing illuminations under white light and pre-defined lighting patterns. Compared to their work, our dataset not only includes OLAT and white light illuminations but also provides color gradients and images captured under environment lights emulated by our lightstage. Moreover, our OLAT data is more fine-grained, with $331$ illuminations compared to $40$ in ReNé and $142$ in OpenIllumination.

In the realm of human-centric lightstage captures, a majority of publicly available datasets focus only on the face, with ICT-3DRFE \cite{ma2007rapidacquisition, stratou2011effectofillumination}, Dynamic OLAT \cite{zhang2021neuralvideoportraitrelighting} and Goliath-4 \cite{saito2024rgca, chen2024urhand} belonging to this list. Note that the latter also provides relightable hand captures. ICT-3DRFE aims to enable experimentation on relightable facial expressions and only provides a 3D face model with the intrinsics necessary for relighting. Dynamic OLAT and Goliath-4 focus on capturing white light and grouped OLAT illuminations of moving subjects, limiting their use outside the dynamic setting. The dataset most similar to ours is Ultrastage \cite{zhou2023relightable}, which contains static relightable captures of $100$ subjects performing different actions under a comparable number of views as in \textit{HumanOLAT}. However, Ultrastage only provides three illuminations---consisting of white light and color gradients---for each capture, limiting their application in methods relying on detailed illumination data. In comparison to these works, we focus on recording multi-view images of the full human body under a multitude of illuminations, including white light, color gradients, environment maps and fine-grained OLATs, for broad-scale applications. 
\section{Baseline Experiments}
\label{sec:Baselines}

This section evaluates several relighting baseline methods using our \textit{HumanOLAT} dataset. We first assess 3DGS-based relighting methods designed for OLAT illumination (Sec.~\ref{subsec:baselineOLAT}). Then, we test our dataset on the illumination harmonization task (Sec.~\ref{subsec:Harmonization}). We also provide an evaluation of Wang et al.~\cite{Wang2024IntrinsicAvatar} in App.~\ref{app:intrinsic_avatar}. These evaluations aim to demonstrate the effectiveness and versatility of our dataset in various relighting scenarios. We present some results in the main paper and refer to App.~\ref{app:additional_results} for more results. 

\subsection{Inverse Rendering from OLAT Illuminations}
\label{subsec:baselineOLAT}

We evaluate the performance of four recent Gaussian-based inverse rendering methods that utilize OLAT illumination from our dataset: PRT-Gaussians \cite{zhang2024prtgaussian}, $GS^3$ \cite{bi2024gs3}, RNG \cite{Fan2025RNG}, and BiGS \cite{zhenyuan2024bigs}. To facilitate efficient training, we downsample the images to 1K resolution and evenly select $100$ lights from $32$ cameras, resulting in ${\approx}3000$ training frames. The remaining cameras and lights are reserved for validation. Since these methods assume static objects, we apply motion compensation as detailed in Section \ref{subsec: motion compensation} to all frames and empirically brighten them by a factor of 10 to strengthen the training signal. We conduct baseline comparisons using the source code provided by the authors in all cases. We evaluate the quantitative results using average PSNR, LPIPS, and SSIM metrics for six representative captures from our dataset, as shown in \cref{tab:prt_gs3_results}. These six captures were chosen to provide a balanced representation of the dataset's diversity. Out of the tested methods, $GS^3$ performs the best overall both qualitatively and numerically. 

We present qualitative results of the tested methods in Fig.~\ref{fig:prt_gs3_qualitative}. While PRT-Gaussian captures some aspects of light transfer, it struggles to accurately reconstruct geometry from the OLAT frames, leading to poor quality and ghosting artifacts. In contrast, $GS^{3}$, RNG \cite{Fan2025RNG}, and BiGS \cite{zhenyuan2024bigs} show superior performance, achieving better geometry and more plausible relighting results. However, as illustrated in Fig.~\ref{fig:prt_gs3_qualitative}, the renderings remain somewhat blurry and display noticeable artifacts in the hand and face regions. These issues might stem from an insufficient number of Gaussians. Although we initialize the method with $300$K sampled points, many are culled during training, resulting in final reconstructions with only approximately $50$K{--}$150$K Gaussians. While $GS^{3}$ achieves the best overall performance, even here complex lighting effects like specular highlights and sharp shadows are not effectively captured.

In summary, while the evaluated methods perform well on simple object-centric scenes, they face challenges with the complex human-centric scenarios in our dataset. Therefore, \textit{HumanOLAT} serves as a crucial benchmark for evaluating and stress testing object-centric relighting methods. 

\subsection{Illumination Harmonization}
\label{subsec:Harmonization}

\begin{figure*}[htb]

\includegraphics[trim=0cm 0cm 0cm 0cm, clip, width=1.0\textwidth]{figures/qualitative_gs_relight_fig.png}
\centering
\caption{Qualitative results for $GS^3$~\cite{bi2024gs3}, PRT-Gaussian~\cite{zhang2024prtgaussian}, RNG~\cite{Fan2025RNG} and BiGS~\cite{zhenyuan2024bigs} under novel views and illuminations.}
\label{fig:prt_gs3_qualitative} 

\vspace{15pt} 

\includegraphics[trim=1.5cm 1.2cm 1.5cm 0.5cm, clip, width=1.0\textwidth]{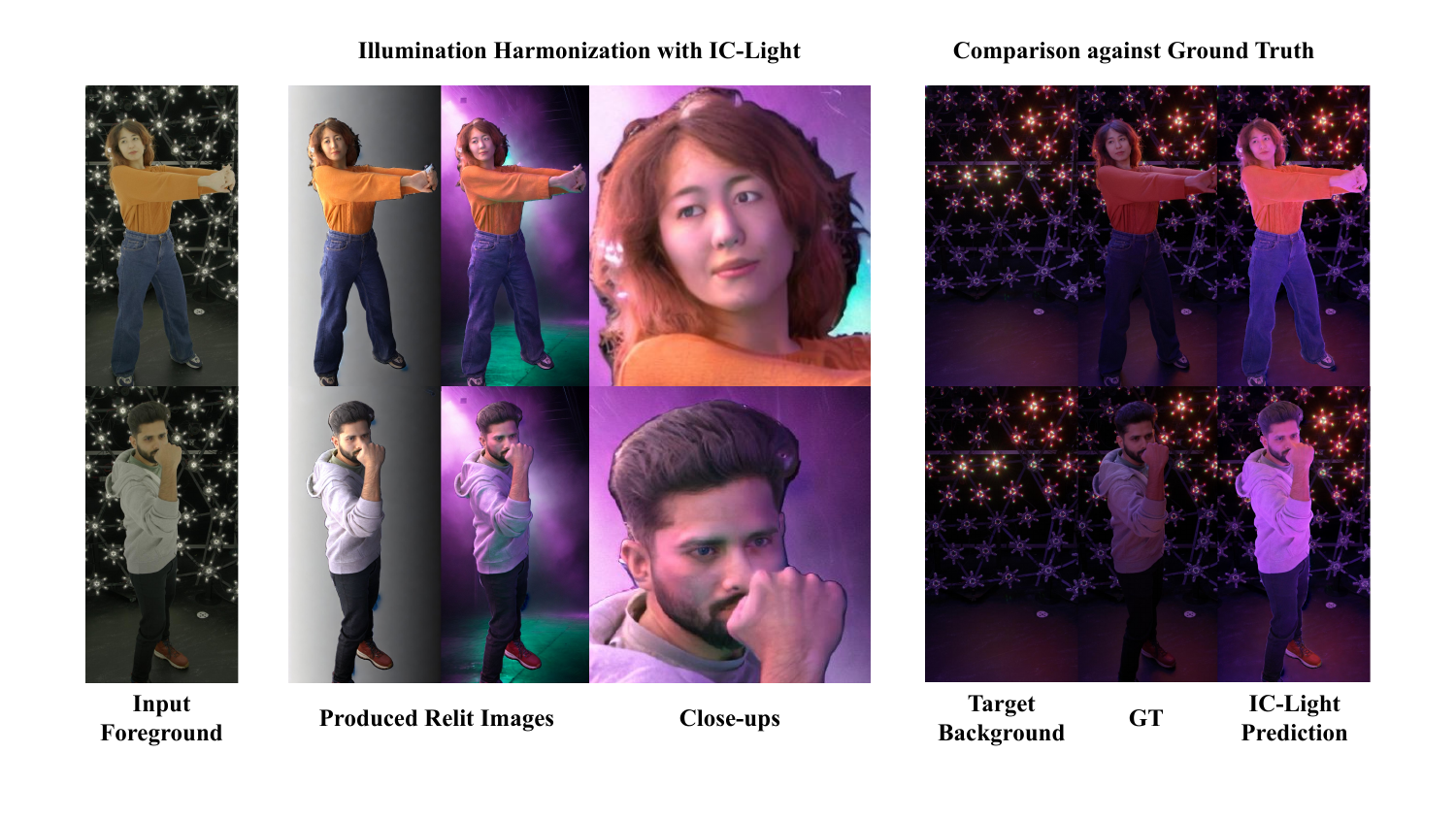}
\centering
\caption{Illumination harmonization with IC-Light~\cite{zhang2025scaling} on \textit{HumanOLAT}. Given an input image (left), we show both relighting results for background images sampled from the selection provided in IC-Light~\cite{zhang2025scaling} (middle) and for a background image of the empty lightstage (right). For latter, we also provide the actual expected illumination of the target given the light emitted by the lightstage.}
\label{fig:ic-light_qualitative}
\end{figure*}

To demonstrate potential applications of our dataset in illumination harmonization, we perform a qualitative survey for the state-of-the-art work IC-Light~\cite{zhang2025scaling} on our data. To evaluate their method, we use the officially released model with our masked foreground images captured under white light and relight using both our background captures and the sample backgrounds provided by IC-Light for six subjects. Qualitative representative results are depicted in Fig.~\ref{fig:ic-light_qualitative}. For detailed quantitative results, please refer to App.~\ref{app:additional_results_quantitative}.

We observe that while the method can produce generally plausible relighting results, it struggles with photorealistic complex light transport effects, such as subsurface scattering, and consistently fails to preserve crucial facial details like eye color and mouth shape. Notably, IC-Light tends to focus on relighting based on normal direction and ambient occlusion, with self-shadows from elements like the head, arms, or clothing rarely observed.  

We attribute these limitations primarily to the training data used by IC-Light, which relies on a mix of publicly available data, including augmented in-the-wild images, synthetically generated multi-illumination data, and lightstage captures for faces and objects, totaling approximately $10^7$ frames. Additionally, they include an unspecified internal OLAT dataset with $2 \cdot 10^4$ appearances. To our knowledge, the method does not utilize significant full-body OLAT data for training, which limits its effectiveness on datasets like ours.

\begin{table} 
\centering
\resizebox{\columnwidth}{!}
{
\begin{tabular}{ c c c c}\toprule
 \multicolumn{1}{c}{Method}& \multicolumn{1}{c}{PSNR $\uparrow$} & \multicolumn{1}{c}{LPIPS $\downarrow$} & \multicolumn{1}{c}{SSIM $\uparrow$} \\ \midrule
  \multicolumn{1}{l}{PRT-Gaussian \cite{zhang2024prtgaussian}}  & $24.06$ & $0.212$ & $0.810$  \\
\multicolumn{1}{l}{$\textbf{\textit{GS}}^{\textbf{3}}$ \cite{bi2024gs3}} & $\textbf{30.04}$ & $\textbf{0.152}$ & $0.892$   \\
\multicolumn{1}{l}{RNG \cite{Fan2025RNG}} & $27.38$ & $0.139$ & $0.905$   \\
\multicolumn{1}{l}{BiGS \cite{zhenyuan2024bigs}} & $26.72$ & $0.201$ & $\textbf{0.936}$  \\\bottomrule
\end{tabular}
}
\caption{Quantitative comparison between the performance of PRT-Gaussian, $GS^3$, RNG and BiGS on OLAT data from the \textit{HumanOLAT} dataset. Results measure the quality of generated images under both novel view and novel illumination, averaged across six subjects with a randomly chosen pose.} 
\vspace{-6mm}
\label{tab:prt_gs3_results}
\end{table}
\section{Discussion and Conclusion}\label{sec:Discussion_Conclusion} 

\vspace{-2mm}
\paragraph{Open Challenges}
Due to complex light transport caused by intricate geometry and diverse materials, photo-realistic depictions of full-body humans under novel illumination remains a challenging problem for many human-centric~\cite{chen2024meshavatarlearninghighqualitytriangular, li2024animatable, xu2023relightableanimatableneuralavatar, lin2023relightableanimatableneuralavatars, Wang2024IntrinsicAvatar, Zheng2024PhysAavatar, iqbal2022ranarelightablearticulatedneural, Luvizon2024relightneuralactor} and general~\cite{kanamori2018relightinghumans, ji2022geometry, He2024DifFRelight, gau2023R3DG, shi2023gir, liang2023gsir, chen2024gi, du2024gsID, wu2024deferredgs, jiang2023gaussianshader, zhu2024gsror, liang2024gusir, zhang2024prtgaussian, zhenyuan2024bigs, kuang2024olat, bi2024gs3, yong2025omgopacitymattersmaterial, dihlmann2024subsurface} relighting approaches. While historically many works rely on data with limited lighting information, various recent works show that captures recorded in a lightstage---in particular, OLAT illuminations---provide suitable ground truth for developing photo-realistic relighting techniques~\cite{kim2024switchlight, pandey2021total, chen2024urhand, saito2024rgca, li2024uravatar, WangARXIV2025fullbodyrgca}. \textit{HumanOLAT} is the first dataset to make such data publicly available, opening the associated avenues of research to the broader community.

\vspace{-5mm}
\paragraph{Conclusion}

This paper introduces \textit{HumanOLAT}, the first publicly available dataset to provide extensive full-body multi-view captures of humans under multiple illuminations, including white light, environment maps, color gradients and fine-grained OLATs. Further, we test our proposed dataset with state-of-the-art baselines for relighting under novel illuminations and novel views. The results demonstrate that while current methods achieve plausible results, they do not capture fully the complex lighting effects characteristic to human bodies. We believe the proposed dataset forms a valuable basis for developing and benchmarking future general and human-centric relighting methods. 

\vspace{-5mm}
\paragraph{Acknowledgement} This research was supported by NVIDIA.\EoA

{
    \small
    \bibliographystyle{ieeenat_fullname}
    \bibliography{main.bib}
}

\clearpage
\maketitlesupplementary

\appendix

This supplement discusses the ethical concerns in App.~\ref{app:ethical_concerns}, provides  additional results 
in App.~\ref{app:additional_results} and samples of the dataset 
in App.~\ref{app:additional_samples}; and describes how we obtain OpenPose annotations and SMPL-X shape parameters in App.~\ref{app:openpose_smplx}.

\vspace{-1mm}

\section{Ethical Concerns}
\label{app:ethical_concerns} 
Every subject was informed about the targeted use for the collected data and consented to making their captures publicly available for scientific purposes. Moreover, we recognize that our dataset could be used in the development of nefarious methods aiming to produce misleading media. To combat potential misuse, researchers are required to explicitly apply for access to \textit{HumanOLAT} and describe their intended use of the dataset. 

\vspace{-1mm}

\section{Additional Results}
\label{app:additional_results} 

\subsection{Additional Quantitative Results}
\label{app:additional_results_quantitative} 

\cref{tab:prt_gs3_results_detailed} provides the detailed per-subject results of the averaged quantitative evaluation of PRT-Gaussian~\cite{zhang2024prtgaussian},  $GS^3$~\cite{bi2024gs3}, RNG \cite{Fan2025RNG} and BiGS \cite{zhenyuan2024bigs} found in \cref{tab:prt_gs3_results}.
Moreover, to supplement the qualitative results shown in \cref{fig:ic-light_qualitative}, we show quantitative results for illumination harmonization against a target background using IC-Light \cite{zhang2025scaling} in \cref{tab:ic_light_quantiative}.  

\subsection{Additional Qualitative Results}
\label{app:additional_results_qualitative} 

\begin{figure*}[htb]

\includegraphics[width=\textwidth]{figures_sup/qualitative_gs_relight_2_fig.png}
\centering
\caption{Additional qualitative results for $GS^3$~\cite{bi2024gs3}, PRT-Gaussian~\cite{zhang2024prtgaussian}, RNG~\cite{Fan2025RNG} and BiGS~\cite{zhenyuan2024bigs} as presented in~\cref{fig:prt_gs3_qualitative}.}
\label{fig:prt_gs3_qualitative_2}

\includegraphics[trim=1.5cm 1cm 1.5cm 0cm, clip, width=1\textwidth]{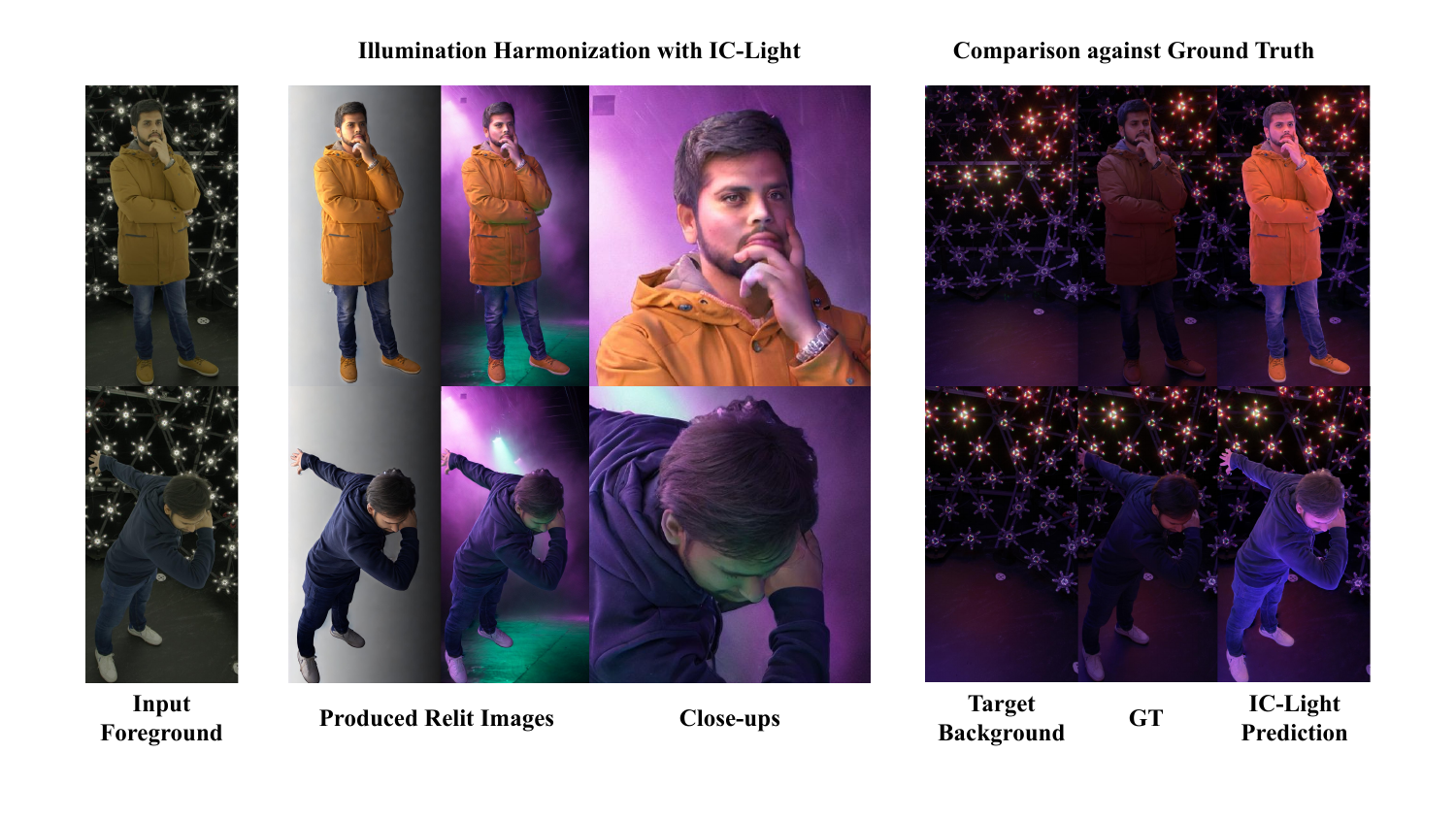}
\centering
\caption{Additional qualitative illustrations for  illumination harmonization with IC-Light~\cite{zhang2025scaling}.}
\label{fig:ic-light_qualitative_2}
\end{figure*}

We provide additional qualitative results for the evaluations presented in \cref{sec:Baselines} in \cref{fig:prt_gs3_qualitative_2} and \cref{fig:ic-light_qualitative_2}.

\begin{table} 

\centering
\resizebox{\columnwidth}{!}
{
\begin{tabular}{ c c c c c}\toprule
 \multicolumn{1}{c}{Method} & \multicolumn{1}{c}{Subject} & \multicolumn{1}{c}{PSNR $\uparrow$} & \multicolumn{1}{c}{LPIPS $\downarrow$} & \multicolumn{1}{c}{SSIM $\uparrow$} \\ \midrule
  \multicolumn{1}{l}{PRT-Gaussian \cite{zhang2024prtgaussian}} &  & $22.64$ & $0.237$ & $0.778$  \\
\multicolumn{1}{l}{$GS^{3}$ \cite{bi2024gs3}} & C003 & $28.44$ & $0.172$ & $0.876$   \\
\multicolumn{1}{l}{RNG \cite{Fan2025RNG}} &  POSE\_00 &  $26.55$ & $0.157$ &  $0.893$   \\
\multicolumn{1}{l}{BiGS \cite{zhenyuan2024bigs}} & & $25.06$ & $0.237$ & $0.924$    \\\midrule
\multicolumn{1}{l}{PRT-Gaussian \cite{zhang2024prtgaussian}} &  & $25.15$ & $0.250$ & $0.794$  \\
\multicolumn{1}{l}{$GS^{3}$ \cite{bi2024gs3}} & C006 &$30.64$ & $0.180$ & $0.876$   \\ 
\multicolumn{1}{l}{RNG \cite{Fan2025RNG}} & POSE\_00  & $28.17$& $0.1518$ & $0.892$   \\
\multicolumn{1}{l}{BiGS \cite{zhenyuan2024bigs}} & & $26.08$ & $0.254$ & $0.914$   \\\midrule
\multicolumn{1}{l}{PRT-Gaussian \cite{zhang2024prtgaussian}} &  & $27.51$ & $0.203$ & $0.842$  \\
\multicolumn{1}{l}{$GS^{3}$ \cite{bi2024gs3}} & C010 & $32.66$ & $0.151$ & $0.906$   \\ 
\multicolumn{1}{l}{RNG \cite{Fan2025RNG}} &  POSE\_01 & $29.43$ & $0.135$ & $0.907$   \\
\multicolumn{1}{l}{BiGS \cite{zhenyuan2024bigs}} & & $30.37$ & $0.177$ & $0.944$    \\\midrule
\multicolumn{1}{l}{PRT-Gaussian \cite{zhang2024prtgaussian}} &  & $23.44$ & $0.185$ & $0.830$  \\
\multicolumn{1}{l}{$GS^{3}$ \cite{bi2024gs3}} & C028 & $30.13$ & $0.131$ & $0.904$   \\ 
\multicolumn{1}{l}{RNG \cite{Fan2025RNG}} & POSE\_01 & $27.34$ & $0.122$ & $0.921$   \\
\multicolumn{1}{l}{BiGS \cite{zhenyuan2024bigs}} &  & $27.05$ & $0.160$ & $0.952$    \\\midrule
\multicolumn{1}{l}{PRT-Gaussian \cite{zhang2024prtgaussian}} & & $24.95$ & $0.182$ & $0.825$  \\
\multicolumn{1}{l}{$GS^{3}$ \cite{bi2024gs3}} & C048 & $30.70$ & $0.137$ & $0.897$   \\
\multicolumn{1}{l}{RNG \cite{Fan2025RNG}} & POSE\_00& $28.80$ & $0.122$ & $0.914$   \\
\multicolumn{1}{l}{BiGS \cite{zhenyuan2024bigs}} & & $28.31$ & $0.169$ & $0.942$    \\\midrule
\multicolumn{1}{l}{PRT-Gaussian \cite{zhang2024prtgaussian}} & & $20.69$ & $0.212$ & $0.792$  \\
\multicolumn{1}{l}{$GS^{3}$ \cite{bi2024gs3}} & C058 & $27.68$ & $0.141$ & $0.894$   \\ 
\multicolumn{1}{l}{RNG \cite{Fan2025RNG}} & POSE\_00 & $24.01$ & $0.146$ & $0.904$ \\
\multicolumn{1}{l}{BiGS \cite{zhenyuan2024bigs}} & & $23.42$ & $0.210$ & $0.942$    \\
 \bottomrule
\end{tabular}
}
\caption{Quantitative per-subject relighting results for OLAT-based relighting methonds.}

\label{tab:prt_gs3_results_detailed}
\end{table}

\begin{table}[h!]
\vspace{-3mm}
\centering
\setlength{\tabcolsep}{5pt}  
\renewcommand{\arraystretch}{1.1}  
\begin{tabular}{@{}lccc@{}}\toprule
Subject & PSNR $\uparrow$ & LPIPS $\downarrow$ & SSIM $\uparrow$ \\ \midrule
C003    & 15.28 & 0.232 & 0.541 \\
C006    & 15.33 & 0.252 & 0.558 \\
C010    & 17.57 & 0.213 & 0.570 \\
C028    & 15.29 & 0.280 & 0.523 \\
C048    & 16.15 & 0.268 & 0.538 \\
C058    & 14.60 & 0.278 & 0.546 \\ \bottomrule
\end{tabular}
\vspace{-2mm}
\caption{Quantitative results for IC-Light~\cite{zhang2025scaling} for six representative subjects.}
\vspace{-10pt}
\label{tab:ic_light_quantiative} 
\end{table}

\subsection{Evaluation on IntrinsicAvatar}
\label{app:intrinsic_avatar}

\begin{figure}[h]
\includegraphics[trim=0cm 0cm 0cm 0cm, clip, width=0.48\textwidth]{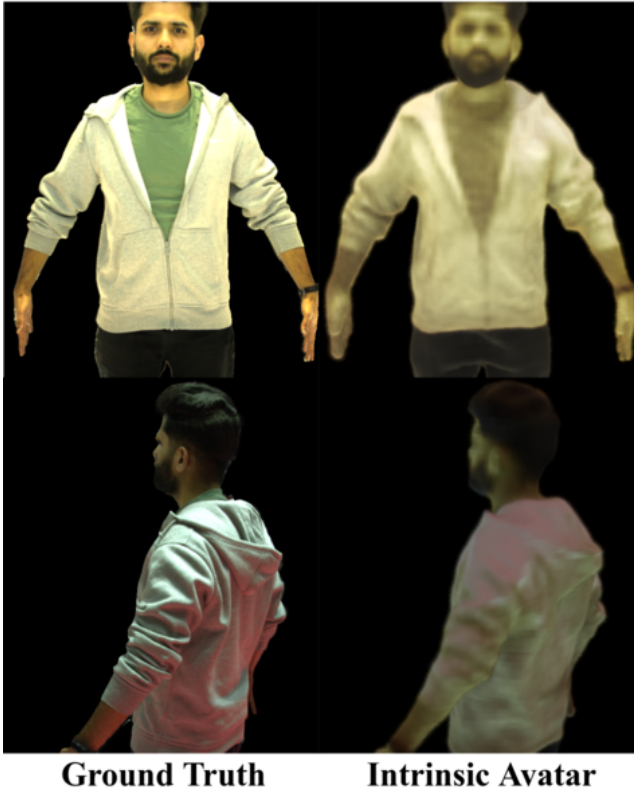}
\centering

\caption{Qualitative and quantitative results for our adapted implementation of IntrinsicAvatar~\cite{Wang2024IntrinsicAvatar}.} 

\label{fig:ia_qualitative}
\end{figure}

In addition to the evaluations presented in \Cref{sec:Baselines}, we adapt IntrinsicAvatar~\cite{Wang2024IntrinsicAvatar} to our setting and train it on one subject from our dataset. We fix one of three poses and supply known lighting conditions during training. Qualitative and quantitative results are shown in Fig.~\ref{fig:ia_qualitative}. While our adapted implementation can learn some aspects of light transfer, it struggles with estimating detailed geometry and produces strongly blurred results.

\vspace{-1mm}

\section{Additional Samples of the Dataset}
\label{app:additional_samples} 

\begin{figure}[ht]
\includegraphics[trim=0cm 0cm 0cm 0cm, clip, width=1.0\columnwidth]{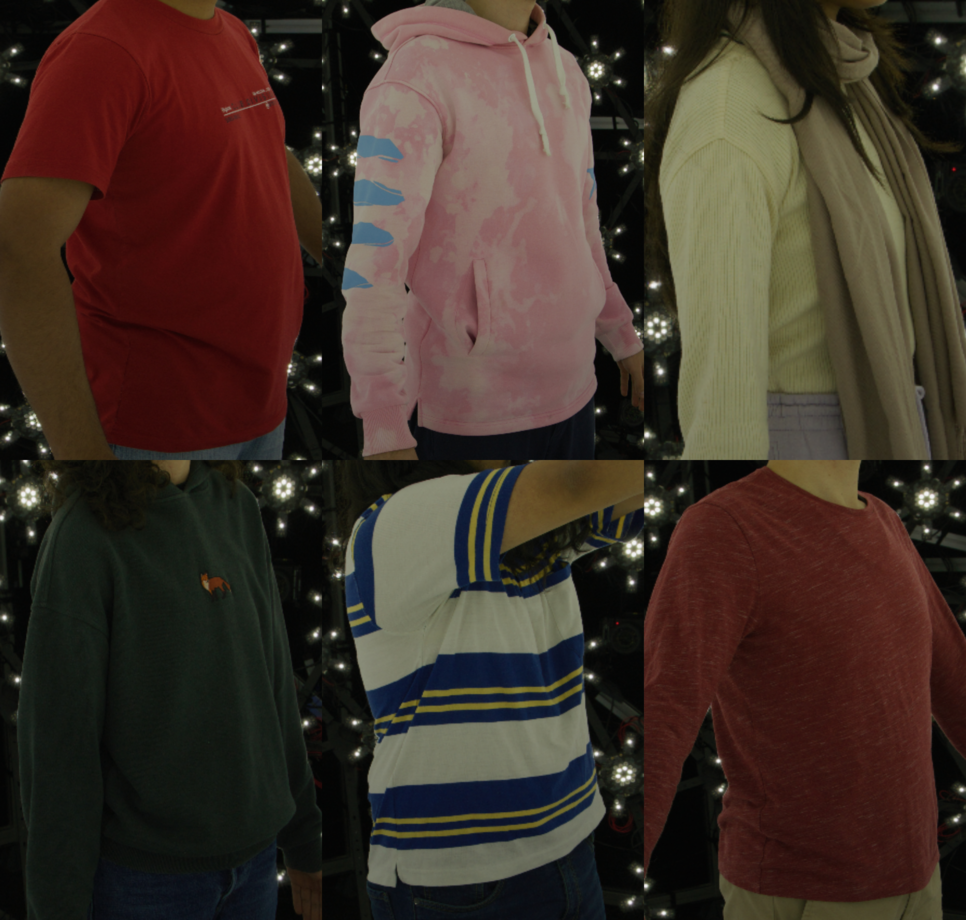}
\centering

\caption{Additional samples showing the variety of clothing in \textit{HumanOLAT}.}

\label{fig:clothing_samples}
\end{figure}

\begin{figure*}[h!]
\includegraphics[trim=0.8cm 1.5cm 0.8cm 1.2cm, clip, width=0.7\textwidth]{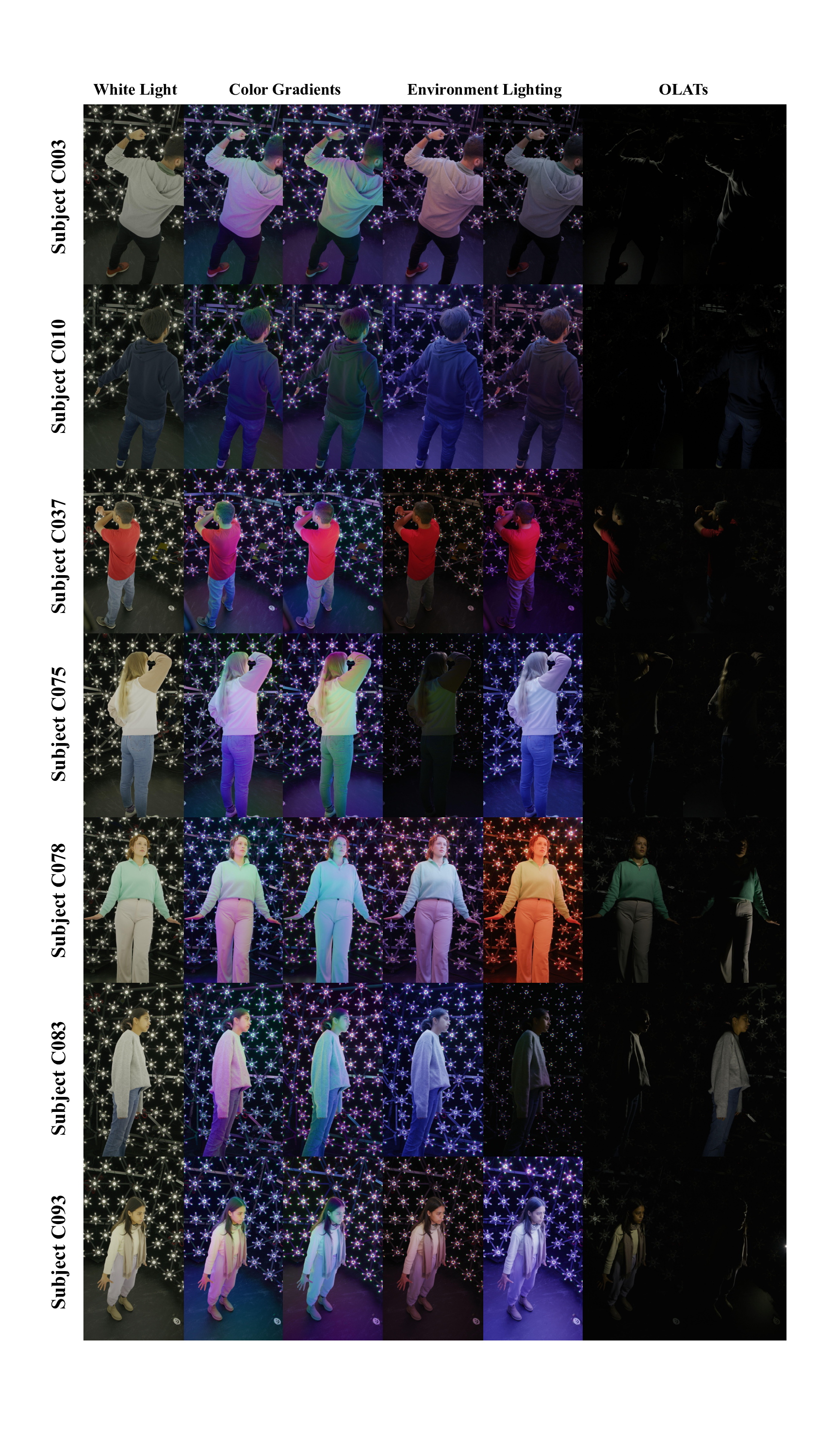}
\centering
\caption{Additional visualizations of samples from \textit{HumanOLAT}. Here, we select seven subjects under a randomly chosen pose and camera view and depict images captured under white light, color gradient and a randomly chosen environment and OLAT illuminations.}
\vspace{-0.2in}
\label{fig:large_data_samples}
\end{figure*}

We show additional samples of captured images from the proposed dataset in \cref{fig:large_data_samples}. Additional samples regarding the clothing variety in our dataset can be found in \cref{fig:clothing_samples}.

\vspace{-1mm}

\section{OpenPose and SMPL-X Annotations} 
\label{app:openpose_smplx} 

\begin{figure}[ht]
\includegraphics[trim=0cm 17.65cm 0cm 0cm, clip, width=0.48\textwidth]{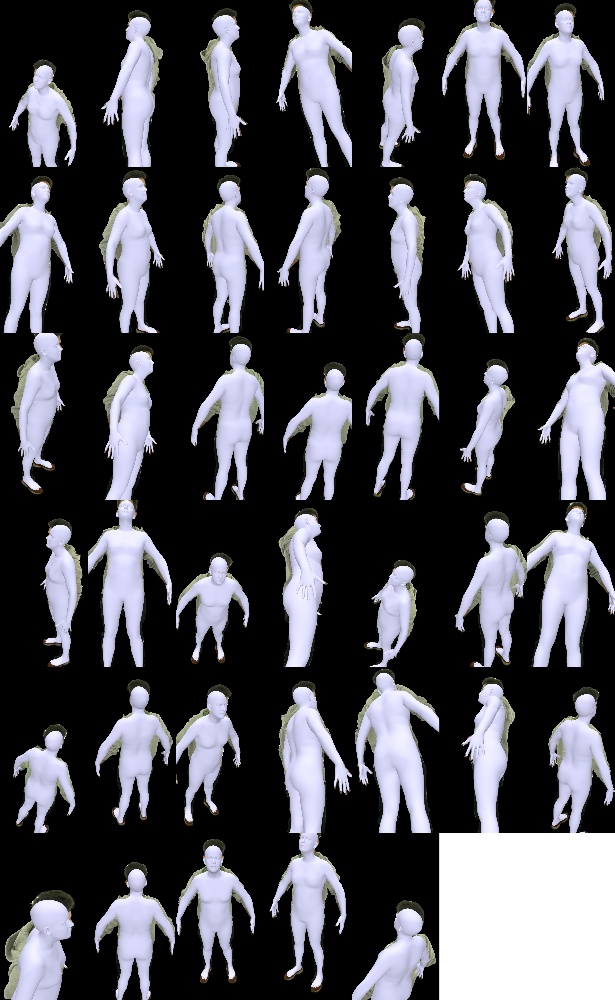}
\vspace{-0.2in}

\caption{Visualization of SMPL-X poses estimated using OpenPose~\cite{cao2019openpose} and EasyMocap \cite{easymocap}.}
\vspace{-0.1in}

\label{fig:smplx}
\end{figure}

We offer pose annotations and SMPL-X~\cite{SMPL-X:2019} parameters estimated using OpenPose~\cite{cao2019openpose} and EasyMocap~\cite{easymocap}, respectively. For the poses, we utilize the recommended pre-built OpenPose Windows binary to generate annotations for all white-light illumination frames. Following, SMPL-X shape parameters are regressed from these annotations using the \texttt{mv1p.py} script provided by EasyMocap. We use the default settings defined by EasyMocap and set the body and gender arguments to \texttt{bodyhandface} and \texttt{neutral}, respectively. 
See \cref{fig:smplx} for a visualization of the SMPL-X estimations.

\end{document}